\documentclass[letterpaper, 10 pt, conference]{ieeeconf}  

\IEEEoverridecommandlockouts                              
\usepackage{comment}
\usepackage{amsmath}
\usepackage{amssymb}
\usepackage{mathtools}
\usepackage{booktabs}
\usepackage{subfigure}
\usepackage{subcaption}  
\usepackage{caption} 
\usepackage{color}
\usepackage[pdftex]{graphicx}
\usepackage{amsfonts} 
\usepackage{booktabs}%
\usepackage{array}
\usepackage{caption}
\usepackage{subcaption}
\usepackage{comment}
\usepackage[pdftex]{graphicx}
\usepackage{dblfloatfix}
\usepackage{comment}
\usepackage{cite}
\usepackage{amsfonts} 
\usepackage[skip=2pt,font=scriptsize]{caption}
\usepackage{amsmath}
\usepackage{hyperref}

\title{\LARGE \bf
Design and Evaluation of a UGV-Based Robotic Platform for Precision Soil Moisture Remote Sensing
}

\author{Ilektra Tsimpidi, Ilias Tevetzidis, Vidya Sumathy, and George Nikolakopoulos 
\thanks{Robotics and AI Team, Department of Computer Science, Electrical and Space Engineering, Luleå University of Technology, Luleå, Sweden}%
}
\begin{document}
\maketitle
\thispagestyle{empty}
\pagestyle{empty}
\begin{abstract}
This extended abstract presents the design and evaluation of AgriOne, an automated unmanned ground vehicle (UGV) platform for high precision sensing of soil moisture in large agricultural fields. The developed robotic system is equipped with a volumetric water content (VWC) sensor mounted on a robotic manipulator and utilizes a surface-aware data collection framework to ensure accurate measurements in heterogeneous terrains. The framework identifies and removes invalid data points where the sensor fails to penetrate the soil, ensuring data reliability. Multiple field experiments were conducted to validate the platform's performance, while the obtained results demonstrate the efficacy of the AgriOne robot in real-time data acquisition, reducing the need for permanent sensors and labor-intensive methods.
\end{abstract}
\section{INTRODUCTION}

Autonomous robotic platforms for precision remote sensing is crucial in advancing robotics research, offering autonomous measuring of soil parameters and collecting data on-site, which are essential for optimizing resource management, and enabling data-driven decision-making across a range of industries, including agriculture. Real-time soil moisture monitoring is essential for optimizing irrigation, enhancing water-use efficiency, and improving crop yields, making it a key area of research in precision agriculture and autonomous farming systems.

In the related literature, extensive research has been conducted across various aspects of this domain, like design of different sensing and sampling sensors, sampling methodologies, and challenges, as highlighted in the following review papers~\cite{li2021soil,yu_2021review,rasheed2022soil,abdulraheem2023advancement,silvero2023sensing,mane2024advancements}, offering a consolidated view of the current state of the field and guiding future directions. Furthermore, the study in~\cite{liu2023artificial} proposes an AI-based analysis method to improve the accuracy of actively heated fiber Bragg grating (AH-FBG) soil moisture measurements, demonstrating that cover conditions, such as bare soil, grass, and biochar, significantly influence results, with artificial neural networks (ANN) effectively reducing measurement errors. The application of the EM38-MK electromagnetic induction device for spatial and vertical mapping of soil electrical conductivity (ECa) to estimate soil salinity and moisture content at multiple depths, utilizing geospatial techniques is investigated in \cite{ratshiedana2023determination}. Another related approach is the use of an integrated robotic soil sensor platform, as presented in~\cite{campbe_ll2021portable}, which employs a ROSbot 2.0 Pro wheeled mobile robot with a Robot Operating System (ROS)-based software stack to autonomously perform geo-referenced ECa measurements. The work~\cite{pulido_2020kriging} introduces an autonomous mobile robot equipped with a non-contact soil moisture sensor that dynamically maps soil moisture and selects optimal sampling locations using a Kriging Variance-based exploration strategy, incorporating Poisson-distributed sensor data for improved accuracy and efficiency in soil moisture modeling.

Drawing on insights from current research and considering existing challenges, this study introduces the AgriOne, an autonomous soil moisture measurement robot leveraging a surface aware data collection framework to achieve precise and efficient soil moisture assessments, thereby minimizing reliance on permanent sensors and reducing associated costs and labor. The major contributions of the work are:
\begin{itemize}
    \item A novel UGV- based robotic platform, the AgriOne, equipped with an advanced volumetric water content sensor mounted on a robotic manipulator.
    \item Development of an innovative on-site surface aware data collection framework for measurements in heterogeneous terrains. 
    \item Real-time experimental assessments and data acquisition with AgriOne robot.
 \end{itemize}
 
\section{METHODOLOGY}
\begin{figure}
      \centering
\includegraphics[width=0.8\linewidth]{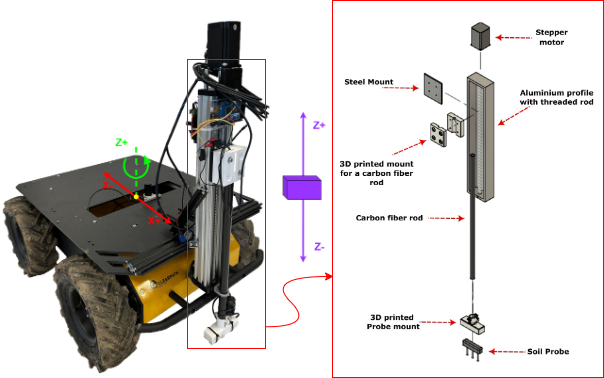}
      \caption{AgriOne robot integrated with Meter TEROS 12 soil sensor used for soil moisture measurement and the schematic view of the linear actuator probe.}
      \label{fig1}
  \end{figure}
\subsection{Design of the robotic platform - AgriOne}
AgriOne, as shown in Fig. \ref{fig1}, is a fully equipped autonomous robotic ground vehicle capable of traversing in uneven terrains, to accurately measure soil temperature, water content and conductivity, together with precise positioning. The robot is based on the commercially available Husky A200 UGV from Clearpath Robotics \cite{clearpath_husky} and it is designed and built by integrating different sensors like Meter TEROS 12 soil sensor \cite{teros_manual}, RTK GPS and an Intel NUC on-board computer, to the UGV platform. The TEROS 12 soil sensor applies a 70-MHz signal to its needles, measuring the charging time proportional to dielectric permittivity and VWC, with the microprocessor outputting a raw value ($RAW$). The soil sensor is mounted on the UGV using a custom linear actuator with an aluminum base and a carbon fiber rod, driven by a stepper motor controlled by an Arduino and stepper motor driver, as shown in Fig. \ref{fig1}. The stepper motor is activated by the Arduino, which receives commands from the Intel NUC companion computer responsible for controlling the motion of the manipulator while collecting all the data. The flow chart in Fig. \ref{fig2} depicts the overall architecture of the proposed methodology. The central unit microcontroller manages communication and control between system components like the Intel NUC, Meter TEROS 12 sensor (via SDI-12), and the manipulator for environmental interactions.

\subsection{Surface aware data collection framework}
\begin{figure}
      \centering
      \includegraphics[width=\linewidth]{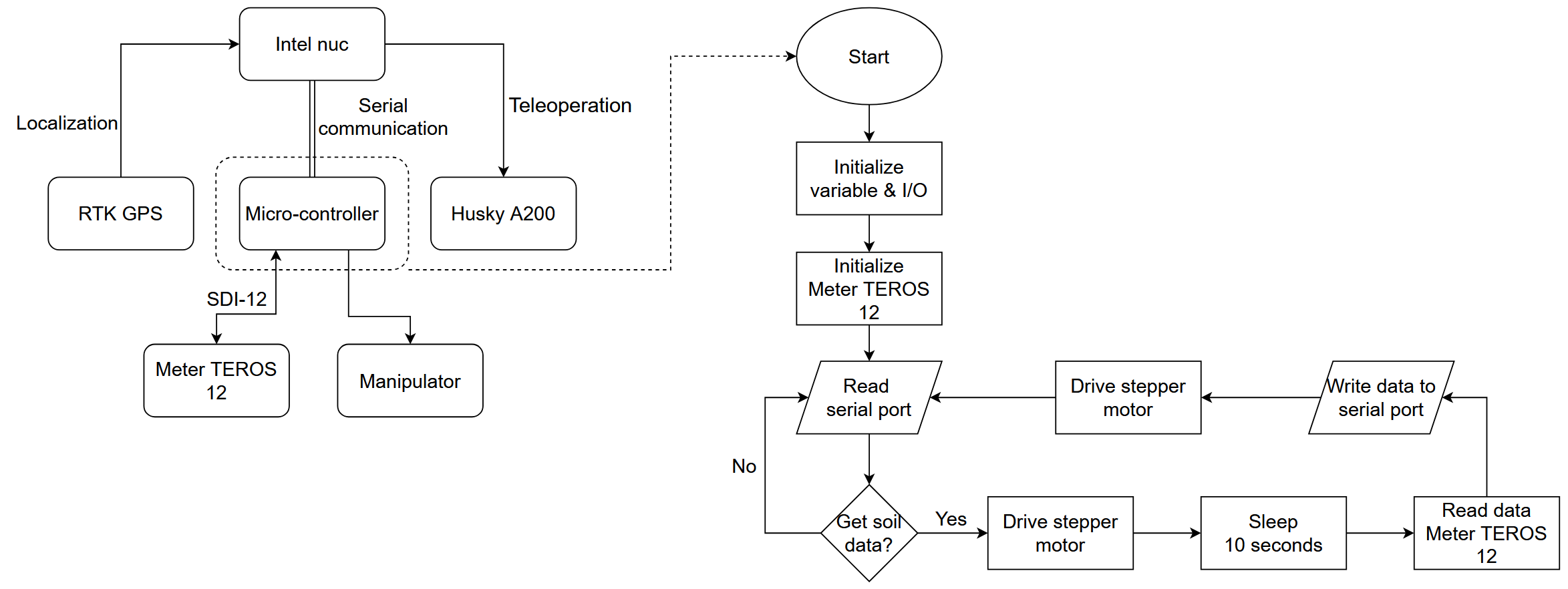}
      \caption{The flowchart depicting the surface aware data collection framework algorithm, based on which AgriOne robot measure and collect soil moisture data.}
      \label{fig2}
  \end{figure}

The Surface Aware Data Collection framework for the AgriOne is designed to automate the validity of the soil data while navigating various terrains. The proposed data collection framework continuously reads soil data, checks its validity, and controls a stepper motor for repeated measurements. This is implemented in the microcontroller, as depicted by the right half of the flow chart shown in Fig. \ref{fig2}. The sensor output ${RAW}$ is converted to VWC by the calibration Eq. \eqref{eq1} \cite{teros_manual},
\begin{equation}\label{eq1}
    \theta (m^{3}/m^{-3}) = 3.879 \times 10^{-4}\times RAW-0.6956
\end{equation} 
\section{Field Experiment}
\begin{figure}
    \centering
    \subfigure[]{\includegraphics[width=0.2\textwidth]{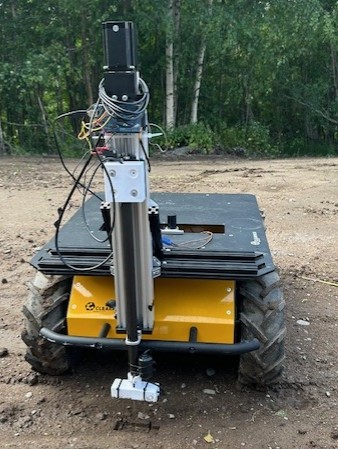}} 
    \subfigure[]{\includegraphics[width=0.2\textwidth]{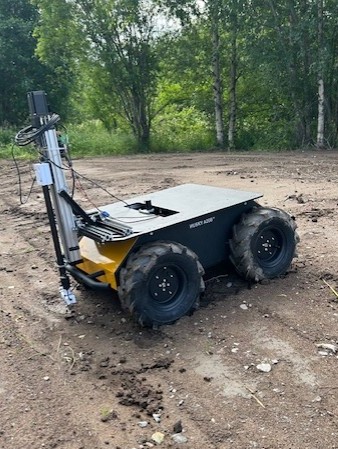}} 
    \caption{(a) Front-view (b) Side-view of the AgriOne robot during soil moisture measurement experiments}
\label{fig:experiment}
\end{figure}

The field experiments were conducted to evaluate the developed AgriOne robot and verify the proposed data collection framework in real environments. The experiments were conducted in an open area characterized by flat terrain with a central slight hill, free from grass and other vegetation, and predominantly covered in soil, as seen in Fig. \ref{fig:experiment}. Measurement points were selected randomly and spaced approximately one meter apart, distributed across both sunny and shaded regions and at a minimum distance of one meter from the trees present along one edge of the test area. The soil sensor, attached to the linear actuator, based on the control command from the companion computer penetrated the soil at a depth of $0.05 m$.
Based on the surface aware data collection framework, any points where the sensor didn't penetrate the soil correctly were promptly identified and removed, ensuring the reliability of our data. The collected data was saved in a rosbag file and extracted via MATLAB software. 
\section{RESULTS}
\begin{figure}
    \centering
    \includegraphics[width=\linewidth]{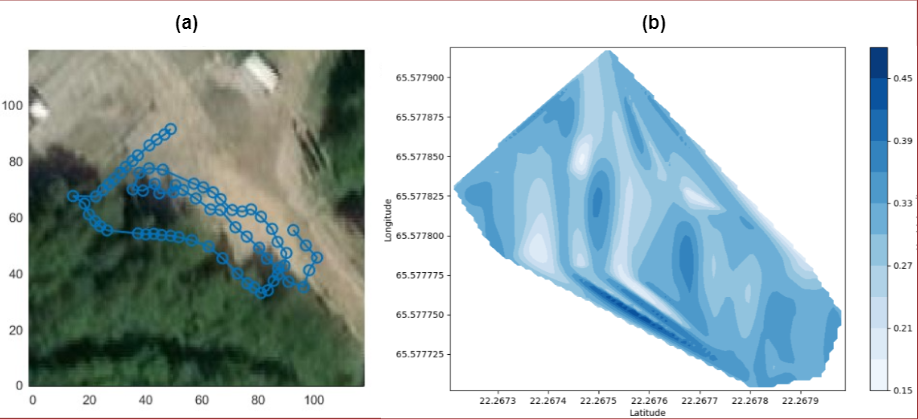}
    \caption{Experiment results: (a) Satellite imagery of the experiment area and the data collection points along the robots's trajectory (b) spatial distribution of soil moisture  across the geographic area based on the collected data samples.}
    \label{fig:representation}
\end{figure}

The experiment was carried out for $57$ minutes covering an area of almost $380m^{2}$.
During the experiment, 95 data collection points were chosen in the test arena, ensuring each one was accurately measured. Out of which, the data from 70 points were collected based on the proposed framework. The results from the data collection experiment are shown in Fig. \ref{fig:representation}, which shows the data points at which the soil moisture was measured and the geospatial distribution of the volumetric water content in the area. Although each point has measurements of its temperature, electrical conductivity, volumetric water content, and geographical coordinates, for this initial experiment, we focus only on the volumetric water content data. The video of the experiment can be found in \url{https://youtu.be/Hu-aDm6tIOg}.
\section{Conclusion}
This extended abstract proposes the design and development of a UGV based robot for soil volumetric water content measurement and a surface aware data collection framework for accurate data collection from the geographic area based on the motion of the linear actuator on which the sensor is attached. The experiment result validates the workability of the AgriOne robot and reliability of the framework for data collection. Future work includes, automated point selection based on data collection and modeling of volumetric water content based on geographic features of the area.
\addtolength{\textheight}{-12cm}   





\bibliographystyle{IEEEtran}
\bibliography{main}
\end{document}